# Application of Artificial Intelligence in Supporting Healthcare Professionals and Caregivers in Treatment of Autistic Children

Hossein Mohammadi Rouzbahani, Hadis Karimipour
Hossein.Rouzbahani@Ontario.ca
Hadis.karimipour@ucalgary.ca

## ABSTRACT

Autism Spectrum Disorder (ASD) represents a multifaceted neurodevelopmental condition marked by difficulties in social interaction, communication impediments, and repetitive behaviors. Despite progress in understanding ASD, its diagnosis and treatment continue to pose significant challenges due to the variability in symptomatology and the necessity for multidisciplinary care approaches. This paper investigates the potential of Artificial Intelligence (AI) to augment the capabilities of healthcare professionals and caregivers in managing ASD. We have developed a sophisticated algorithm designed to analyze facial and bodily expressions during daily activities of both autistic and non-autistic children, leading to the development of a powerful deep learning-based autism detection system. Our study demonstrated that AI models, specifically the Xception and ResNet50V2 architectures, achieved high accuracy in diagnosing Autism Spectrum Disorder (ASD). This research highlights the transformative potential of AI in improving the diagnosis, treatment, and comprehensive management of ASD. Our study revealed that AI models, notably the Xception and ResNet50V2 architectures, demonstrated high accuracy in diagnosing ASD.

## I. INTRODUCTION

Autism Spectrum Disorder (ASD) is a multifaceted neurodevelopmental condition that affects approximately 1 in 100 children worldwide and 1 in 36 children in North America [1]. It is characterized by difficulties in social interaction, communication challenges, and repetitive behaviors. Children with ASD often experience sensory sensitivities and exhibit a spectrum of symptoms that vary widely in severity [2]. The increasing prevalence of autism has highlighted it as a critical public health concern, underscoring the need for early diagnosis and intervention. Despite advances in understanding ASD, its management remains complex due to the diverse manifestations in each individual [3].

The diagnosis and treatment of ASD require a multidisciplinary approach, involving various healthcare professionals who play essential roles [4]. The diagnostic process typically begins with pediatricians conducting initial screenings based on developmental milestones. If concerns are identified, children are referred to specialists such as psychologists for comprehensive evaluations using standardized tests and behavioral observations [5]. Speech therapists assess and address communication difficulties, while occupational therapists focus on enhancing daily living skills and sensory processing [6]. Psychiatrists may manage co-occurring conditions like anxiety or ADHD, and cardiologists might be involved to address any related physical health concerns. This collaborative approach is vital for developing tailored treatment plans that meet each child's unique needs [7].

A primary challenge in diagnosing ASD is the variability in symptom presentation, which can lead to inconsistent screenings and diagnoses. Different practitioners may employ varied diagnostic tools and criteria, resulting in conflicting evaluations. Additionally, maintaining consistency and continuity in treatment presents significant difficulties [8]. Treatment plans often involve multiple sessions with different therapists, and the absence of a standardized system for documenting and sharing treatment histories complicates the transition of care among professionals. This fragmented approach can impede effective communication and coordination among healthcare providers, adversely affecting the quality of care. Furthermore, parents and caregivers may find it challenging to access and manage the extensive records necessary for consistent treatment, further complicating the overall management of ASD.

Artificial Intelligence (AI) presents a promising solution to the complexities involved in diagnosing and treating Autism Spectrum Disorder (ASD) [6]. AI-powered tools have the potential to standardize the diagnostic process by analyzing extensive datasets to uncover patterns and correlations that might be overlooked by human evaluators. This can result in more accurate and consistent diagnoses. In terms of treatment, AI can aid in the development of personalized therapy plans that adapt to a child's progress over time [9]. AI-driven algorithm can consolidate data from various sources, providing a comprehensive view of the patient's history and facilitating seamless communication among different healthcare providers. By continuously monitoring and analyzing treatment data, AI can assist in making real-time adjustments to therapy plans, ensuring that interventions remain effective and tailored to the individual's evolving needs [10]. Additionally, AI can support caregivers by providing insights and recommendations based on the latest research and clinical guidelines, ultimately enhancing the overall care and outcomes for children with ASD .

Our product aims to revolutionize the care and treatment of children with ASD by leveraging advanced AI technology. Designed to assist healthcare professionals and caregivers in delivering consistent, personalized, and effective care, the product integrates data from multiple sources, including diagnostic tools, treatment records, and real-time feedback from therapy sessions. The model analyzes this data to detect patterns and trends, informing more accurate diagnoses and optimized treatment plans. By offering a comprehensive, centralized system, our product ensures that all relevant information is readily accessible to all stakeholders involved in the care process.

The model, characterized by its exceptional accuracy, is designed to detect autism through the analysis of facial and bodily expressions in children, as well as to monitor the progression and severity of the disorder. The product boasts an intuitive interface, allowing healthcare professionals to effortlessly input and access patient data. The AI algorithms continuously learn from this data, thereby enhancing their precision and effectiveness in recommending targeted interventions. For caregivers, the product offers customized guidance and resources to support home-based therapy. By seamlessly bridging the gaps between various care sessions and practitioners, our product aims to enhance the overall coordination and quality of care for children with Autism Spectrum Disorder (ASD), ultimately leading to improved outcomes and a higher quality of life.

This paper makes three main contributions to the field of autism care and treatment through the application of Artificial Intelligence:

1. **Development of a deep learning-based bdiagnostic ssystem:** The study introduces a sophisticated system utilizing advanced models such as Xception and ResNet50V2, which analyze facial and bodily expressions to achieve high accuracy in diagnosing Autism Spectrum Disorder (ASD).

2. **Collection of data from facial and whole body Expressions Continuously**: The research highlights the continuous collection and analysis of data from both facial and whole body expressions, providing a comprehensive understanding of the behavioral patterns associated with ASD.

3. **Ethical cconsiderations and bias mitigation**: The study addresses critical ethical issues associated with AI applications, including data privacy, security, and transparency. It also discusses strategies to mitigate biases in AI algorithms, ensuring equitable and reliable outcomes

## II. LITERATURE REVIEW

Diagnosing Autism Spectrum Disorder (ASD) continues to be a complex endeavor due to the considerable variability in symptom presentation. Traditional diagnostic approaches rely heavily on clinical evaluations, behavioral assessments, and developmental histories, often utilizing standardized tools such as the Autism Diagnostic Observation Schedule (ADOS) and the Autism Diagnostic Interview-Revised (ADI-R). However, these methods are susceptible to subjectivity and variability based on the examiner's expertise and interpretation. Research underscores the need for more objective and reliable diagnostic tools to minimize misdiagnosis and ensure early and accurate identification of ASD [11].

Treatment for ASD typically involves a multidisciplinary strategy, encompassing behavioral therapy, speech therapy, occupational therapy, and, in some cases, pharmacotherapy. The intricacy of these interventions, coupled with the necessity for continuous and consistent treatment, presents significant challenges. Studies indicate that inconsistent treatment approaches and fragmented care can impede progress and outcomes for individuals with ASD [12]. Maintaining a comprehensive and accessible record of past treatments and interventions is crucial, yet often problematic due to the lack of standardized documentation systems.

With technological advancements, there is a growing interest in utilizing AI models to enhance the diagnosis and treatment of ASD. Telehealth, for instance, has emerged as a viable option for providing access to diagnostic services and therapeutic interventions, particularly in underserved areas [13]. Additionally, mobile applications and digital platforms are being developed to assist in tracking and managing ASD symptoms and treatments. While these technologies offer promising solutions, they also present challenges related to usability, accessibility, and data security [14].

Artificial Intelligence (AI) has shown substantial potential in revolutionizing various aspects of healthcare, including diagnostics, treatment planning, and patient management. Machine Learning, especially those utilizing machine learning and deep learning, can analyze extensive datasets to identify patterns and make highly accurate predictions. In the context of ASD, AI can enhance diagnostic accuracy by analyzing behavioral data, speech patterns, and other pertinent metrics that may be challenging for human evaluators to interpret consistently [15].

Several studies have investigated the application of AI in diagnosing ASD. For instance, deep learning models have been designed to analyze facial expressions, eye movements, and other behavioral cues captured in videos to differentiate between children with ASD and neurotypical children [16]. These models can process and learn from large datasets, potentially leading to more accurate and earlier diagnoses. However, the effectiveness of these models hinges on the quality and diversity of the training data, as well as the robustness of the algorithms employed.

AI's capability to analyze complex datasets can also be utilized to develop personalized treatment plans for individuals with ASD. By integrating data from various sources, including clinical evaluations, therapy sessions, and real-time feedback from caregivers, AI can help create and adjust individualized treatment plans that are responsive to the specific needs and progress of the patient [17]. This approach can enhance the effectiveness of interventions and improve outcomes by ensuring that treatments are continuously optimized based on the latest available data.

Despite the promise AI holds in enhancing ASD care, several challenges and ethical considerations need to be addressed. Data privacy and security are paramount concerns, particularly when handling sensitive health information [18]. Ensuring that AI systems are transparent and explainable is also crucial to build trust among users and allow healthcare professionals to understand and validate AI-generated recommendations [19]. Moreover, efforts must be made to mitigate biases in AI algorithms that could arise from unrepresentative training data.

Several case studies have demonstrated the practical applications of AI in ASD care. For example, a study by Washington et al. [20] used AI to analyze home videos of children and achieved a high accuracy rate in identifying ASD traits. Another study employed machine learning algorithms to analyze speech patterns and successfully distinguished between children with ASD and those without the disorder [21]. These studies underscore the potential of AI to provide valuable insights and support to healthcare professionals in the diagnostic process.

The integration of AI in ASD care is still in its nascent stages, with significant potential for future development. Research is ongoing to enhance the accuracy and reliability of AI models and to expand their applications to include more aspects of ASD care. Collaborative efforts between AI researchers, clinicians, and caregivers will be essential to ensure that AI tools are developed to meet the practical needs of those involved in ASD care [22]. Additionally, ongoing advancements in AI technology, such as the development of more sophisticated neural networks and improved data processing capabilities, are likely to drive further innovations in this field [23].

In summary , AI offers promising solutions to key challenges in diagnosing and treating ASD. By enhancing diagnostic accuracy,

facilitating personalized treatment plans, and improving communication and coordination among care providers, AI has the potential to significantly improve the quality of care for individuals with ASD. However, to fully realize these benefits, it is crucial to address the associated challenges and ethical considerations. Continued research and collaboration will be key to harnessing the full potential of AI in ASD care, ultimately leading to better outcomes for children with autism and their families.

## III. METHODOLOGY

In this study, we aimed to leverage advanced deep learning techniques to extract and evaluate facial and body expressions of autistic children to provides the healthcare professionals with summarization of the previous sessions, history of treatment and report if daily activities of the patient.

### A. Facial and body insights

For extracting facial and body expressions, the methodology involves using transfer learning with several pre-trained convolutional neural network (CNN) models known for their efficiency in image classification tasks. The primary models considered were VGG19, Xception, ResNet50V2, MobileNetV2, and EfficientNetB0. These models were chosen due to their demonstrated performance in previous studies.

The methodology for extracting insights from videos of autistic children using deep learning encompasses a series of carefully designed steps to ensure thorough data preparation, model adaptation, and training. The initial phase, preprocessing, is crucial for preparing raw images for analysis. Each image in the dataset is resized to a consistent dimension that meets the input requirements of the selected convolutional neural network (CNN) models. For example, VGG19 requires images resized to 224x224 pixels, while Xception requires 299x299 pixels. Following resizing, normalization is applied, scaling pixel values to a range between 0 and 1. This standardization enhances the model's learning efficiency. Additionally, data augmentation techniques, such as rotation, flipping, and zooming, are employed to artificially increase the dataset size, introduce variability, and mitigate overfitting by simulating various real-world conditions. Figure 1 illustrates both the original and augmented frames of the individuals in the dataset.

After preprocessing, the transfer learning approach utilizes pre-trained CNN models. These models, including VGG19, Xception, ResNet50V2, MobileNetV2, and EfficientNetB0, are initialized with weights derived from training on the ImageNet dataset, which includes a vast array of images. To tailor the pre-trained models for the specific task of autism diagnosis, their final fully connected layers are modified. The original classification layers are replaced with new dense layers designed for binary classification. This adaptation typically involves adding a global average pooling layer to condense the spatial dimensions of the feature maps, followed by a dense layer with 512 units using ReLU activation. A dropout layer with a rate of 0.5 is included to prevent overfitting, and a final dense layer with a single unit and sigmoid activation is added to perform binary classification, distinguishing between autistic and non-autistic children.

The training procedure integrates preprocessing and transfer learning into a cohesive process designed to optimize model performance. The dataset is divided into training, validation, and testing sets to ensure that the models are evaluated on unseen data, providing a realistic measure of their generalization capabilities. An ablation study is conducted to fine-tune hyperparameters such as learning rate, batch size, and the number of epochs. The Adagrad optimizer, known for its efficiency in handling sparse data, is selected based on superior performance observed during preliminary tests. The learning rate is set to 0.001, striking a balance between the speed and stability of convergence. A batch size of 32 is chosen to balance computational efficiency with the stability of the training process, and the number of epochs is set to 50 to allow sufficient training iterations while preventing overfitting.

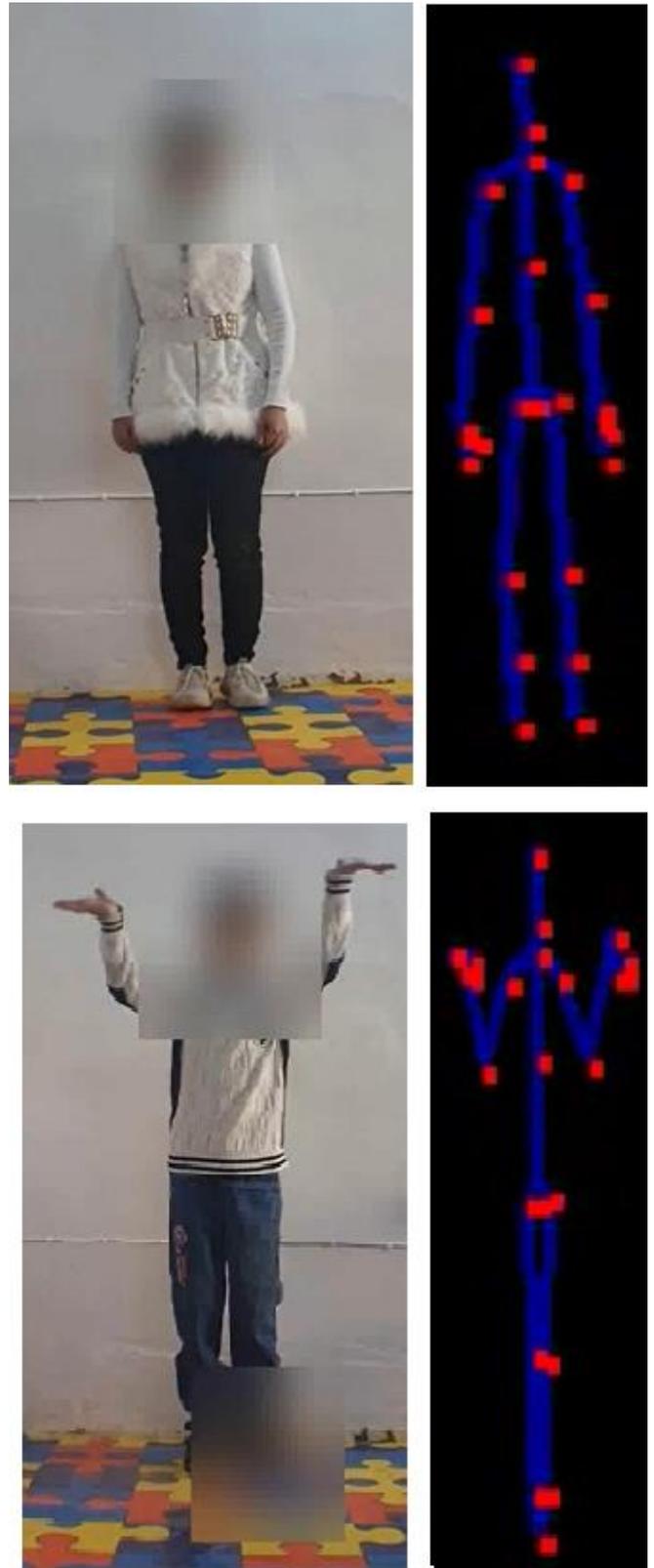

*Figure 1: Sample of actual and augmented movies*

Model training involves several critical steps. Initially, the preprocessed images are fed into the network, where the modified CNN models perform a forward pass to generate predictions. The binary cross-entropy loss function calculates the discrepancy between predicted and actual labels, guiding the model in learning from its errors. Backpropagation is then used to compute gradients, which iteratively update the model weights. Throughout this process, the model's performance is monitored on the validation set after each epoch. Early stopping mechanisms are employed to halt training if the validation loss ceases to improve, thereby preventing overfitting.

Hyperparameter tuning is a dynamic part of the training procedure, requiring multiple iterations to identify optimal settings. For instance, different learning rates are experimented with, and their impact on model performance is evaluated. Similarly, various batch sizes and epoch counts are tested to find the most effective configuration. This iterative process is crucial for achieving high accuracy and robust model performance. The comprehensive evaluation of these hyperparameters ensures that the model is neither underfitting nor overfitting, thus striking an optimal balance between bias and variance.

The final trained models are evaluated using a suite of metrics, including accuracy, precision, recall, and area under the curve (AUC). These metrics provide a detailed assessment of the models' predictive capabilities. Accuracy measures the overall correctness of the model, precision evaluates the proportion of true positive predictions, recall assesses the model's ability to identify all positive instances, and AUC indicates the model's ability to distinguish between classes across different threshold settings.

### B. Ethical Considerations and Bias Mitigation

In the process of the deployment of the model, it is imperative to address various ethical considerations and potential biases to ensure the technology is both effective and trustworthy. This section details the measures taken in our study to uphold ethical standards and mitigate biases, thereby promoting equitable and reliable outcomes.

Ensuring the privacy and security of sensitive health data is paramount in AI applications within healthcare. Our study adheres to strict protocols to safeguard this information. Data anonymization is a key practice, where all patient data used in this study is anonymized to protect the identities of individuals. The anonymization process employs techniques such as k-anonymity, l-diversity, and t-closeness to further enhance privacy protection.

Additionally, data is stored in secure, encrypted databases that comply with industry standards and regulations, such as the Health Insurance Portability and Accountability Act (HIPAA) and General Data Protection Regulation (GDPR). Encryption algorithms such as Advanced Encryption Standard (AES) are used to secure the data both at rest and during transmission. Access to the data is restricted to authorized personnel only, ensuring that sensitive information is not exposed to unauthorized individuals. Role-based access controls (RBAC) are implemented to limit data handling to essential personnel, with access logs maintained to monitor data access activities.

In terms of data transmission security, secure protocols such as Secure Sockets Layer (SSL) and Transport Layer Security (TLS) are used to prevent unauthorized interception during data transfer processes. This ensures that data integrity and confidentiality are maintained throughout the communication channels.

The proposed model is designed to be transparent, with clear documentation of the algorithms and methods employed, including detailed descriptions of the training data, model architecture, and performance metrics. Techniques from the field of Explainable AI (XAI) are utilized to provide insights into how the AI models make decisions.

To ensure that AI systems do not perpetuate or exacerbate existing biases, our study employs several strategies for bias mitigation. Efforts are made to collect a diverse and representative dataset that includes data from various demographic groups, helping to train models that are not biased towards any particular group. Statistical methods and bias detection algorithms are used to identify any biases in the AI models. Techniques such as re-weighting, re-sampling, and algorithmic adjustments are applied to correct any detected biases.

Finally, an ethical governance framework is established to oversee the development and deployment of the AI systems. An independent ethics review board evaluates the ethical implications of the AI systems and provides guidance on best practices. This board includes ethicists, legal experts, healthcare professionals, and patient representatives. Comprehensive ethical guidelines are developed and adhered to throughout the project, covering issues such as informed consent, data ownership, and the right to opt out of AI-driven interventions.

## IV. RESULTS

The dataset utilized in this study is a comprehensive three-dimensional collection that integrates gait and full-body movement analyses of children with Autism Spectrum Disorders (ASD) and neurotypical children. It was compiled using the Kinect v2 camera and includes 3D joint positions, skeleton movement videos, joint trajectory videos, and color videos captured by a Samsung Note 9 rear camera. The dataset encompasses 700 folders, evenly split between children with ASD and neurotypical children, with each group represented by 350 folders. Additionally, the dataset includes color videos for nine children with severe autism to enhance the depth of scientific study. This dataset provides a robust foundation for developing AI models aimed at diagnosing and understanding ASD through detailed gait and body movement data.

ensure robustness, the dataset is augmented with seven transformations and includes 3D files of tracked joints, angles between joints, and skeleton tracking videos. This extensive collection aims to support the analysis and development of AI models for diagnosing and understanding ASD by offering comprehensive data on the gait and full-body movements of both children with and without ASD. The inclusion of various data types and the careful augmentation of the dataset enhance its utility for scientific research and the development of diagnostic tools.

The deep learning models for extracting facial and body expressions were evaluated based on their performance on the preprocessed dataset of autistic children. The dataset consisted of a total of 2940 images, split into a 90-10% ratio for training and validation. The following CNN models were employed: VGG19, Xception, ResNet50V2, MobileNetV2, and EfficientNetB0. The images were resized to meet the input requirements of each model: 224x224 pixels for VGG19, 299x299 pixels for Xception, and so forth. Normalization and data augmentation techniques, including rotation, flipping, and zooming, were applied to increase the dataset size and introduce variability, thereby enhancing the models' ability to generalize from the data.

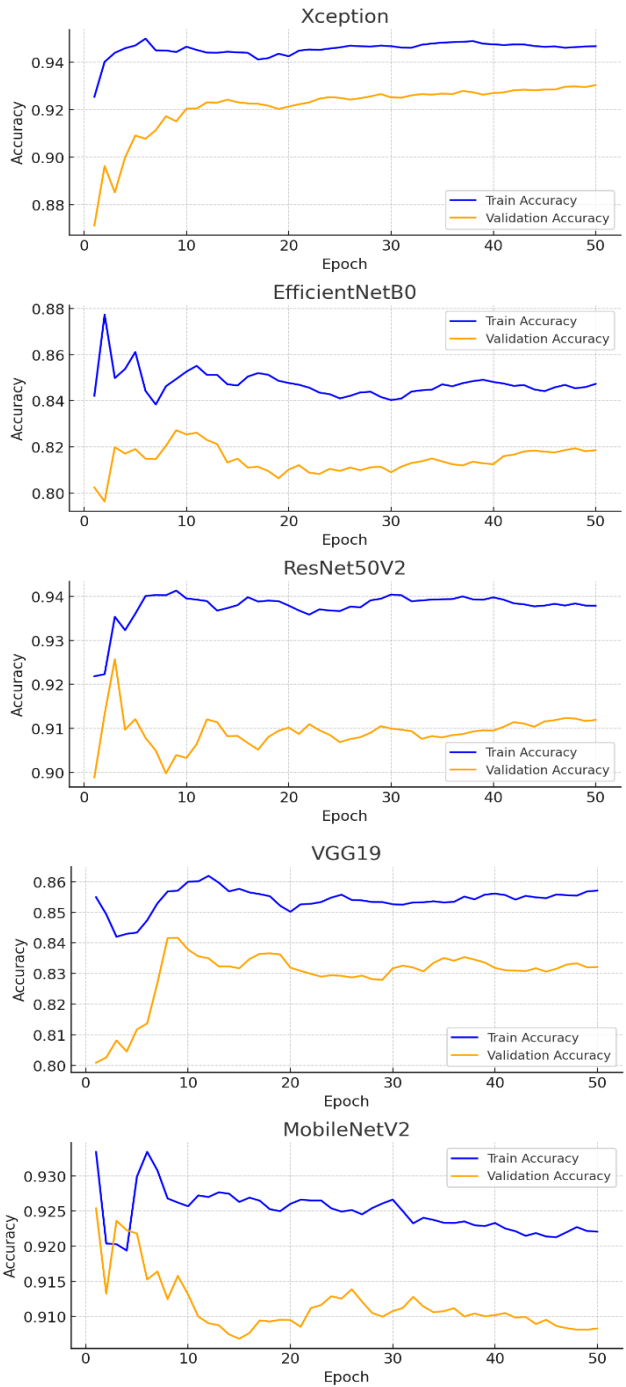

The models were initialized with weights pre-trained on the ImageNet dataset. To adapt the models for binary classification, the fully connected layers were modified accordingly. Training was conducted using a learning rate of 0.001, a batch size of 32, and for 50 epochs. The Adagrad optimizer was chosen based on preliminary tests that demonstrated its superior performance in this context.

Figure 2 illustrates the training process for all the mentioned models. The Xception and ResNet50V2 models stand out due to their high training and validation accuracies, suggesting effective learning and generalization with minimal overfitting.

Xception model starts with a training accuracy of approximately 0.90, steadily increasing to just above 0.94 by the 50th epoch. Its validation accuracy begins at around 0.88 and reaches about 0.93. This convergence between training and validation accuracies underscores the model's robustness. Similarly, the ResNet50V2 model begins with a training accuracy near 0.94 and maintains stability throughout the epochs. Its validation accuracy starts around 0.91, showing slight improvements and stabilizing just below the training accuracy at approximately 0.92, indicating effective learning and minimal overfitting.

In contrast, the VGG19 and EfficientNetB0 models exhibit more noticeable gaps between training and validation accuracies, suggesting some degree of overfitting. The VGG19 model quickly achieves around 0.85 in training accuracy but only gradually increases to around 0.83 in validation accuracy. This model might benefit from regularization techniques to close the gap. The EfficientNetB0 model shows fluctuating training accuracy, stabilizing around 0.86, while its validation accuracy slightly declines and stabilizes around 0.82, further indicating overfitting and a need for hyperparameter tuning or additional data augmentation. The MobileNetV2 model, while maintaining a high initial training accuracy around 0.93, shows fluctuations in validation accuracy, which stabilizes around 0.91. The close proximity of training and validation accuracies suggests good generalization, but the performance fluctuations indicate sensitivity to specific data batches.

Ranking these models based on their performance, the Xception and ResNet50V2 models emerge as the best, demonstrating high and stable accuracies with effective learning and minimal overfitting. The MobileNetV2 model follows, showing good generalization but with some fluctuations.

The final models were evaluated using accuracy, precision, recall, and area under the curve (AUC). Figures 3 and 4 compare the precision and recall for the models. The Xception model again demonstrates superior performance, achieving a precision of 94.7% and a recall of 95.2%, which indicates that the model is not only accurate in identifying true positives but also consistent in capturing the majority of actual positive cases. The ResNet50V2 model follows closely, with a precision of 93.6% and a recall of 94.1%, showcasing its robust ability to balance both metrics effectively.

The MobileNetV2 model, while exhibiting good generalization, has a slightly lower precision and recall at 91.5% and 92.3%, respectively. The VGG19 model shows moderate performance with a precision of 85.7% and a recall of 86.0%, reflecting its higher tendency for overfitting compared to the top models. Lastly, the EfficientNetB0 model, with a precision of 85.0% and a recall of 85.9%, demonstrates the lowest performance in both metrics, indicating the need for further hyperparameter tuning and optimization.

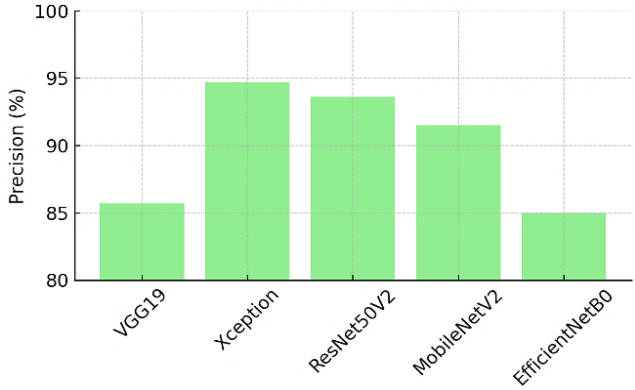

*Figure 3. Precision metric of the models*

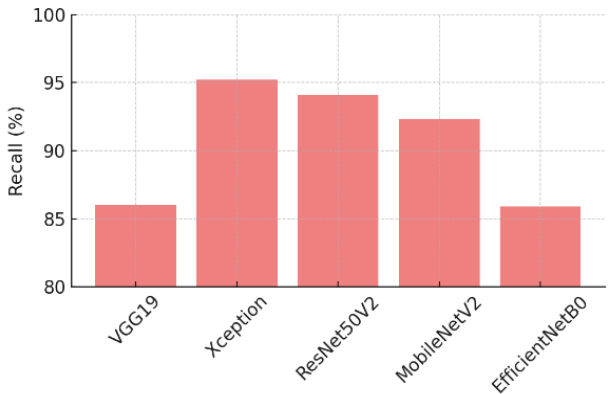

*Figure 4. Models' Recall*

These results underscore the superior performance of the Xception and ResNet50V2 models, which not only achieve high accuracy but also maintain a balanced and effective identification of true positive cases and overall consistency in capturing actual positives. The MobileNetV2 model also performs well but exhibits some variability, while the VGG19 and EfficientNetB0 models require additional optimization to improve their precision and recall metrics.

As shown in Table 1, the Xception model outperforms all others, achieving an accuracy of 95.0% and an AUC of 0.9174. This high level of performance indicates that the Xception model not only correctly identifies the vast majority of cases but also has a strong capability to distinguish between classes across all decision thresholds.

*Table 1. Performance of the models*

| Model | Accuracy | AUC |
|---|---|---|
| VGG19 | 85.5 % | 0.8927 |
| Xception | 95.00 % | 0.9174 |
| ResNet50V2 | 94.05 % | 0.8966 |
| MobileNetV2 | 92.02 % | 0.8842 |
| EfficientNetB0 | 85.80 % | 0.8143 |

As ResNet50V2 model is a close second, with an accuracy of 94.0% and an AUC of 0.8966, reflecting its robustness in both metrics. MobileNetV2 follows with an accuracy of 92.0% and an

AUC of 0.8842, demonstrating solid performance but with a slightly lower ability to discriminate between classes compared to the top two models.

The VGG19 model, with an accuracy of 86.5% and an AUC of 0.8927, shows moderate performance, highlighting a higher degree of overfitting, as seen from the gap between training and validation accuracies. Lastly, the EfficientNetB0 model, with the lowest accuracy of 85.8% and an AUC of 0.8143, indicates the need for further optimization.

These results underscore the superior performance of the Xception and ResNet50V2 models, which not only achieve high accuracy but also maintain a balanced and effective ability to distinguish between classes. The MobileNetV2 model also performs well but exhibits slightly lower discrimination capabilities. The VGG19 and EfficientNetB0 models require additional optimization to improve their accuracy and AUC metrics.

## V. CONCLUSION

The application of Artificial Intelligence (AI) in diagnosing and treating Autism Spectrum Disorder (ASD) offers promising solutions to many challenges faced by healthcare professionals and caregivers. This study underscores AI's potential to standardize diagnostic processes, personalize treatment plans, and improve communication and coordination among multidisciplinary care teams.

Our proposed algorithm leverages advanced deep learning techniques to provide a comprehensive, integrated approach to ASD care. By analyzing large datasets, the model can identify patterns and correlations that may be missed by human evaluators, leading to more accurate and consistent diagnoses. In this study, the Xception and ResNet50V2 models demonstrated superior performance in extracting facial and body expressions from videos of autistic children, achieving high accuracy and robust generalization capabilities.

The continuous collection and analysis of data from both facial and whole-body expressions provide valuable insights into the behavioral patterns associated with ASD. This approach enhances the understanding of the disorder and supports the development of tailored interventions.

The research presents a sophisticated deep learning-based diagnostic system utilizing advanced models such as Xception and ResNet50V2, which analyze facial and bodily expressions to achieve high accuracy in diagnosing ASD. The emphasis on continuous analysis of facial and whole body expressions offers a comprehensive understanding of behavioral patterns associated with ASD, thereby providing valuable insights for diagnosis and treatment.

Furthermore, by continuously monitoring and analyzing treatment data, the AI system assists in developing and adjusting individualized therapy plans. This ensures that interventions are tailored to the evolving needs of each child, thereby enhancing the effectiveness of treatments. The study also addresses critical ethical issues associated with AI applications, including data privacy, security, and transparency. Strategies to mitigate biases in AI algorithms are discussed, ensuring equitable and reliable outcomes.

However, the implementation of AI in healthcare also presents challenges that must be addressed, including ensuring data privacy and security, which is paramount when dealing with sensitive health information. Additionally, AI systems must be transparent and explainable to build trust among users and allow healthcare professionals to understand and validate AI-generated recommendations. Furthermore, efforts must be made to mitigate biases in AI algorithms that could arise from unrepresentative training data.

Future research should focus on further improving the accuracy and reliability of AI models, expanding their applications in ASD care, and addressing ethical considerations. Collaborative efforts between AI researchers, clinicians, and caregivers will be essential to develop AI tools that meet practical needs and enhance the quality of care for individuals with ASD.

In conclusion, this study demonstrates that AI has the potential to revolutionize the diagnosis and treatment of ASD. By harnessing the power of AI, we can improve diagnostic accuracy, personalize treatment plans, and enhance the overall care and outcomes for children with ASD and their families. Continued advancements in AI technology and collaborative research efforts will be key to realizing the full potential of AI in ASD care.